\documentclass[runningheads]{llncs}
\usepackage{graphicx}
\usepackage{graphicx}
\usepackage{amsmath}
\usepackage{amssymb}
\usepackage{booktabs}
\usepackage{cite}
\usepackage{amsmath,amssymb,amsfonts}
\usepackage{graphicx}
\usepackage{textcomp}
\usepackage{xcolor}
\usepackage{multirow}
\usepackage{algorithm}
\usepackage{algorithmic}
\usepackage[pagebackref=true,breaklinks=true,colorlinks,bookmarks=false]{hyperref}
\newtheorem{subtask}{Subtask}

\usepackage{tikz}
\usepackage{comment}
\usepackage{amsmath,amssymb} % define this before the line numbering.
\usepackage{color}

\usepackage[accsupp]{axessibility}  % Improves PDF readability for those with disabilities.

\begin{document}
% \linenumbers
\pagestyle{headings}
\mainmatter
\def\ECCVSubNumber{5880}  % Insert your submission number here

\title{Self-Supervised Interactive Object Segmentation Through a Singulation-and-Grasping Approach} % Replace with your title

\begin{comment}
\titlerunning{ECCV-22 submission ID 5880} 
\authorrunning{ECCV-22 submission ID 5880} 
\author{Anonymous ECCV submission}
\institute{Paper ID 5880}
\end{comment}
%******************

% \begin{comment}
\titlerunning{SaG Interactive Segmentation}

\author{Houjian Yu \and Changhyun Choi}
\authorrunning{H. Yu and C. Choi.}
\institute{Department of Electrical and Computer Engineering\\University of Minnesota, Twin Cities\\
% , Minneapolis, MN 55455, USA
\email{\{yu000487,cchoi\}@umn.edu}}
% \end{comment}
%******************
\maketitle

\begin{abstract}
Instance segmentation with unseen objects is a challenging problem in unstructured environments. To solve this problem, we propose a robot learning approach to actively interact with novel objects and collect each object's training label for further fine-tuning to improve the segmentation model performance, while avoiding the time-consuming process of manually labeling a dataset. 
% The Singulation-and-Grasping (SaG) policy is trained through end-to-end reinforcement learning. 
Given a cluttered pile of objects, our approach chooses pushing and grasping motions to break the clutter and conducts object-agnostic grasping for which the Singulation-and-Grasping (SaG) policy takes as input the visual observations and imperfect segmentation. We decompose the problem into three subtasks: (1) the object singulation subtask aims to separate the objects from each other, which creates more space that alleviates the difficulty of (2) the collision-free grasping subtask; (3) the mask generation subtask obtains the self-labeled ground truth masks by using an optical flow-based binary classifier and motion cue post-processing for transfer learning. Our system achieves $70\%$ singulation success rate in simulated cluttered scenes. The interactive segmentation of our system achieves $87.8\%$, $73.9\%$, and $69.3\%$ average precision for toy blocks, YCB objects in simulation, and real-world novel objects, respectively, which outperforms the compared baselines. Please refer to our project page for more information: \url{https://z.umn.edu/sag-interactive-segmentation}.
\end{abstract}

\section{Introduction}

Instance segmentation is one of the most informative inputs to visual-based robot manipulation systems. It greatly accelerates the robotic learning process while improves the motion efficiency for target-oriented tasks \cite{xu2021efficient,kurenkov2020visuomotor,fang2018multi,liang2021learning}. However, in real cases, robot agents frequently encounter novel objects in unstructured environments, exacerbating the accuracy of object segmentation \cite{9382336,xie2020best}. 
% without a perfect objectness detector trained on corresponding annotations. 
In such a situation, humans often employ multiple interactions with the unknown objects and perceive object segments having consistent motions. This allows us to eventually get familiar with the novel objects and understand their shapes and contours \cite{spelke1990principles}. Our work aims to enable robots to perform the same task. Given an imperfect segmentation model and unseen objects, our robot agent learns to obtain object segment labels in a self-supervised manner via object pushing and grasping interactions and then improves its segmentation model to perceive the novel objects more effectively.

\begin{figure}[t]
\centering
\includegraphics[width=0.6\linewidth]{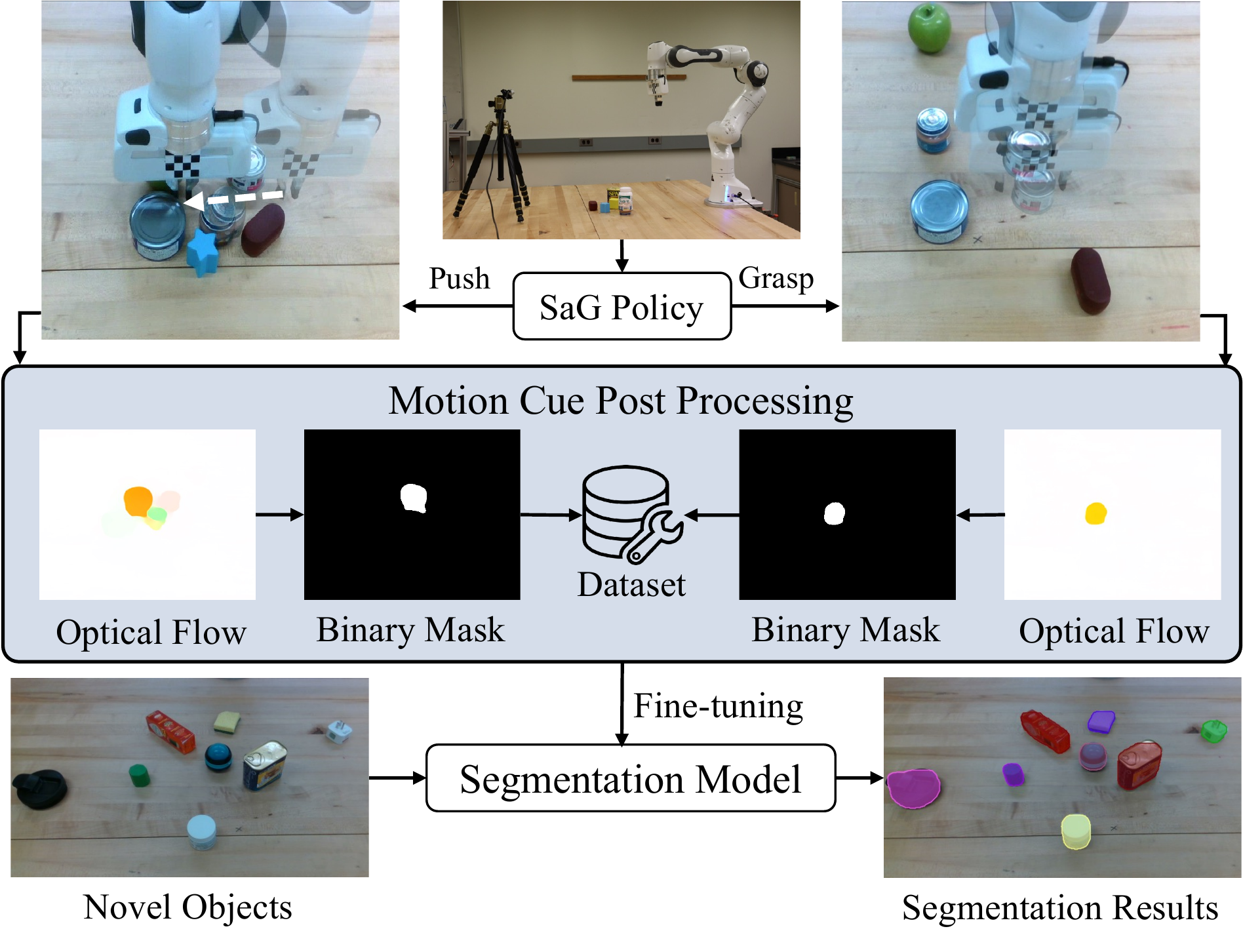}
\caption{The robot agent learns a Singulation-and-Grasping (SaG) policy via deep Q-learning in simulation. We collect the RGB images before and after applying the actions and use coherent motion to create pseudo ground truth masks for the segmentation transfer learning.}
\label{fig1}
\end{figure}

% One of the main challenges for robot manipulation tasks is a lack of ground truth annotations for novel objects. Without objects' spatial arrangement information from perception, the performance on target-oriented grasping and pushing tasks would be undesirably low. 

Classical learning-based segmentation methods require a large amount of human-labeled training annotation, such as ImageNet \cite{russakovsky2015imagenet} and MS COCO dataset \cite{lin2014microsoft}. While these methods have shown generalization to novel objects to some extent, they underperform when objects are out of the distribution of the trained objects. Interactive segmentation approach has taken an orthogonal avenue by actively collecting labels for novel objects using a robotic manipulator. 
% For the existing interactive segmentation methods, 
Pathak et al. uses picking-and-placing  \cite{pathak2018learning} and Eitel et al. adopts pushing for singulation  \cite{eitel2019self} to generate single object location displacement and obtain the ground truth label. However, these methods are limited because the simple frame difference method in~\cite{pathak2018learning} is noisy in label annotations and inefficient when multiple objects move simultaneously due to grasping collisions and failures. The work in \cite{eitel2019self} requires a relatively accurate segmentation method and a large amount of hand-labelled pushing actions to train their push proposal network \cite{eitel2020learning} beforehand and cannot be directly applied to unseen scenes. 

To address the limitations above and obtain high quality object annotations with minimal human intervention, we propose a Singulation-and-Grasping (SaG) pipeline free of laborious manual annotation to improve the segmentation results through robot-object interaction. Fig. \ref{fig1} shows our solution to the problem. The main contributions of our work are as follows:
\begin{itemize}
  \item We train the Singulation-and-Grasping (SaG) policy in an end-to-end learning of a Deep Q-Network (DQN) without human annotations.
  \item We propose a data collection pipeline combining both the pushing and grasping motions to generate high-quality pseudo ground truth masks for unseen objects. The segmentation results after transfer learning show that our method can be used for unseen object segmentation in highly cluttered scenes.
  \item We evaluate our system in a real-world setting without fine-tuning the DQN, which shows the system generalization capability.
\end{itemize}

\section{RELATED WORK}

\noindent\textbf{Interactive Segmentation} Previous works dealing with the interactive perception problem focus on generating interactions with the environment based on objectness hypotheses and obtaining feedback after applying actions to update the segmentation results in recognition, data collection, and pose estimation tasks \cite{bohg2017interactive, fitzpatrick2003first, kenney2009interactive,kuzmivc2010object}. Many methods use a robot manipulator to distinguish one object from the others by applying pre-planned non-prehensile actions to specific object hypothesis \cite{chaudhary2016retrieving, schiebener2014physical, le2017segmenting}. However, the non-prehensile action, such as pushing, for a specific object is challenging in cluttered environments due to inevitable collisions with other surrounding objects. Our work utilizes both pushing and grasping actions to facilitate object isolation from a clutter.

The SE3-Net~\cite{byravan2017se3} learns to segment distinct objects from raw scene point clouds and predicts an object's rigid motion, but it only considers up to 3 objects in a less dense clutter. The closest works to our approach are Pathak et al.~\cite{pathak2018learning} and Eitel et al.~\cite{eitel2019self} that use grasping and pushing, respectively. However, both of them require collision-free interactions to work effectively. \cite{boerdijk2020self} exploits motion cue to differentiate the grasped novel objects from the manipulator and background and gets single object annotation. However, in real cases, the model trained on such data will not reach high performance in heavy clutter. Our Singulation-and-Grasping (SaG) policy manages to solve the collision problem during grasping and even obtains the pseudo ground truth annotations during the singulation phase. %Meanwhile, the object annotations are collected from multi-object scenarios where the domain gap is relatively small between training and testing.

\noindent\textbf{Pushing and Grasping Collaboration} Synergistic behaviors between pushing and grasping have been well explored in \cite{zeng2018learning, huang2021dipn, yang2020deep, deng2019deep, chen2020combining}. The visual pushing for grasping (VPG) \cite{zeng2018learning} provides a model-free deep Q-learning framework to jointly learn pushing and grasping policies, where the pushing action is applied to facilitate future grasps. Both \cite{huang2021dipn} and \cite{yang2020deep} use robust foreground segmentation methods, which track object location through interaction. In such a case, the ground truth transformation for each object can be matched, and the reward from measurements such as border occupancy ratio \cite{deng2019deep} can be designed accordingly. \cite{deng2019deep} and \cite{chen2020combining} conduct grasps by actively exploring and making rearrangements of the environment until the rule-based grasp detect algorithm or the DQN decide whether the goal object is suitable for grasping. Analogous to these methods, the visual system in our work cannot provide robust tracking information before and after the interaction, especially when the objects are previously unseen. In such a case, the reward design for our DQN is much more challenging. Our system instead collects high quality data annotation rather than achieve a simple object removal task. 

\noindent\textbf{Object Singulation} Previous work \cite{eitel2020learning} effectively solves the singulation problem but uses human-labeled pushing actions to train a push proposal network. On the other hand, \cite{hermans2012guided} selects pushing actions to verify if visible edges correspond to proposed object boundaries without learning features. \cite{sarantopoulos2020split} and \cite{kiatos2019robust} focus on the target-oriented object singulation problem, however, it is much more challenging to train a singulation policy that separates all objects than a target-oriented singulation policy that separates only one target object from a clutter. As such, our approach focuses on an object-agnostic singulation problem.

\section{Problem Formulation}
We formulate the interactive object segmentation problem as follows:
\begin{definition}
Given multiple novel objects on a planar surface, the manipulator executes pushing or grasping motion primitives based on the potentially noisy segmentation results (e.g., under- or over-segments). The goal is to improve the segmentation performance via fine-tuning with the data collected during robot-object interactions.
\end{definition}

To solve the interactive data collection problem, we divide the problem into three subtasks:
\begin{subtask}
Given a pile of novel objects, the robot executes objects singulation motions to separate them from each other to increase free space, facilitating grasping actions later. We define this task as the \textbf{object singulation} task.
\end{subtask}
\begin{subtask}
Given a well-singulated scene where the pairwise distances of object segments are above a threshold, the robot grasps and removes objects from the scene. We define this task as the \textbf{collision-free grasping} task.
\end{subtask}

\begin{subtask}
Given the RGB images collected from the previous two subtasks, the binary segmentation masks are generated by using a learned classifier and a motion cue post-processing. We define this task as the \textbf{mask generation} task.
\end{subtask}

\section{Method}

We model the problem as a discrete Markov Decision Process (MDP) as in \cite{zeng2018learning} and \cite{yang2020deep}. Given a state $s_t$, the agent executes an action $a_t$ according to the trained policy $\pi(s_t)$ and obtains the new state $s_{t+1}$ receiving a current reward $R_{a_t}(s_t,s_{t+1})$. The goal of our network is to obtain an action-value function $Q_{\pi}(s_t,a_t)$ that approximates the expected future return for each motion $a_t$. We also introduce the Singulation-and-Grasping (SaG) pipeline, an interactive data collection process, from which objects annotations are self-generated.
% specifically the singulation and collision-free grasping policies

\begin{figure}[t]
\centering
\includegraphics[width=1\textwidth,height=0.65\textwidth]{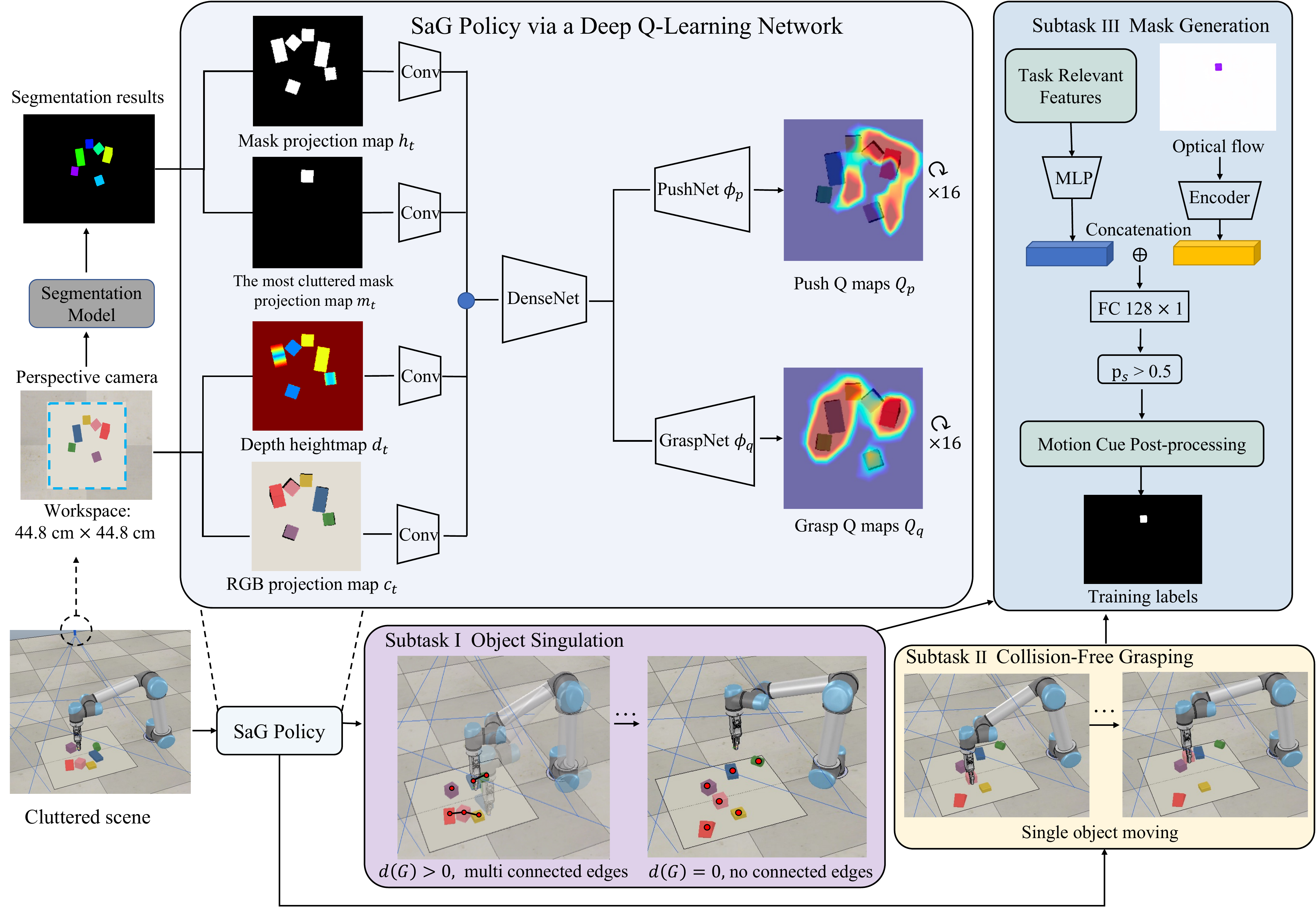}
\caption{\textbf{The SaG pipeline for interactive data collection.} The deep Q-network takes as input the state representation $s_t$, which consists of the orthographically projected RGB-D images ($c_t, d_t$) and object segmentation masks ($h_t, m_t$). The initially cluttered objects are singulated and grasped via the SaG policy $\pi$. Both the scenes of interaction and the task-relevant features are recorded to obtain object segment annotations.}
\label{fig2}
\end{figure}

\subsection{System Overview}
As illustrated in Fig. \ref{fig2}, an RGB-D camera is affixed to the environment to provide visual information of the workspace. The original RGB-D image $I_t$ at time $t$ is first segmented by a segmentation model to provide objectness hypotheses. In this work, we use the UOIS segmentation model \cite{9382336} taking the RGB and depth images to make inferences. By treating each segmented connected component $\{s^1,s^2,...,s^n\}$ as a single object instance, we relabel the segmentation results and get 2D instance center locations $\{c^1,c^2,...,c^m\}$ based on their axis-aligned bounding box coordinates. 

We orthographically project the RGB, depth, segmentation hypotheses, and the most cluttered mask in the gravity direction with known camera parameters to get the color projection map $c_t \in \mathbb{R}^{H\times W\times 3}$, depth heightmap $d_t \in \mathbb{R}^{H\times W\times 1}$, mask projection map $h_t \in \mathbb{R}^{H\times W\times 1}$, and the most cluttered mask projection map $m_t \in \mathbb{R}^{H\times W\times 1}$ (see Section~\ref{sec:method:sag} for the details of $m_t$). The mask projection map $h_t$ introduces the global clutter distribution information to the system, while the most cluttered mask projection map $m_t$ highlights the possible target object that requires most effort to be singulated. This additional input $m_t$ intuitively suggests removing the most cluttered area during singulation. During the grasping stage, $m_t$ is set to an all-ones map. The state is represented by $s_t=(c_t,d_t,h_t,m_t) \in \mathbb{R}^{H\times W\times 6k}$, where we rotate the state representation $k$ times before feeding in the network to reason about multiple orientations for motions. We set $k = 16$ with a fixed step size of $22.5^\circ$ w.r.t. the z-axis. The feature extractor (a two-layer residual network block \cite{he2016deep}) takes $s_t$ as input and further passes to a pre-trained DenseNet-121 \cite{huang2017densely}. The PushNet $\phi_p$ and GraspNet $\phi_g$ finally predict the Q-maps in which each pixel value represents the expected future return if the motion is applied to the pixel location and the corresponding orientation. To maximize the reward, the pushing and grasping motion primitives are executed at the highest Q-value in the Q-maps~\cite{Mnih15nature, zeng2018learning, yang2020deep}.
% This network structure is proved to be effective in \cite{zeng2018learning} and \cite{yang2020deep}.

\subsection{Singulation-and-Grasping Pipeline}
\label{sec:method:sag}
We use the singulation and collision-free grasping motion primitives as the main interaction mechanism. The singulation policy and grasping policy are trained in an multi-stage manner:

\textbf{Stage \uppercase\expandafter{\romannumeral1}: Singulation Only Training.} In this stage, we train the PushNet $\phi_p$ for object singulation. We initially form a densely-cluttered scene where objects are close to each other. During the singulation stage, an undirected-graph structure $G=(V,E)$ is formed for each state $s_t$, where $V :=\{1,...,m\}$, $E\subset V \times V$ and each node $i \in V$ is represented by $c^i$. The edge $E$ is constructed by the Euclidean distance between nodes. When the pairwise distance is under a threshold $p$, an edge connects two nodes. We then find the most cluttered mask from the segmentation hypothesis that corresponds to the largest number of connected edges.

To effectively train the singulation policy via the PushNet $\phi_p$, we need to carefully design a reward function $R_p$. 
Existing target-oriented methods have a strong assumption that a target object can be robustly detected in the course of interactions~\cite{xu2021efficient,  yang2020deep}. 
We relax that assumption since the segmentation hypotheses in $h_t$ is possibly noisy (i.e., over- or under-segments may exist) in the presence of novel objects. In that case, it is challenging to robustly segment/track objects. Instead, we employ a set of surrogate measures. 
% The perception model then provides under-and-over segmentation as a zero-shot inference.
To represent the degree of singulation of the scene, we obtain the graph density value~\cite{coleman1983estimation} as 
\begin{equation}
d(G)=\frac{2|E|}{|V|(|V|-1)}
\end{equation}
where $|E|$ represents the number of the edges and $|V|$ is the number of the vertices. For a cluttered scene, the vertices in the graph are highly connected, and hence the density value $d(G)$ is close to one. In contrast, when objects are well singulated, the density value $d(G)$ is close to zero.

% In contrast, our segmentation model cannot track the pushed objects, which makes designing the reward more challenging. 
A good singulation motion is supposed to create an end-effector trajectory across the segmentation masks but also separate the under-segmented clutter, resulting in a non-decreasing number of object masks. Additionally, effective motions should increase average pairwise distance between objects and decrease the graph density $d(G)$. We also consider a two-dimensional multivariate Gaussian distribution $\mathcal{N}$ fitted to the center locations of the object segments $c^i$. The determinant of the covariance matrix $\Sigma$ of $\mathcal{N}$ indicates the sparsity of the spatial distribution. Therefore, we design the pushing reward function as:

\begin{equation}
R_{p}= \begin{cases} $-0.5$, &\text{$d(G)$ increases} \\ 0.25, &\text{pushing passes mask $h_t$ and}\\ &\text{$|\{c^1,...,c^m\}|_{t+1}$ non-decreases}  \\ 0.5, &\text{$a_d$ or $a_{var}$ increases} \\1.0, &\text{$d(G)$ decreases or $|\Sigma|$ increases} \end{cases}
\end{equation}
where $a_d$ and $a_{var}$ indicate the average and variance of the pairwise center location distance, respectively. $|\{c^1,...,c^m\}|_{t+1}$ represents the number of instance masks at time $t+1$. $|\Sigma|$ represents the determinant of the covariance matrix $\Sigma$ of $\mathcal{N}$. 

\textbf{Stage \uppercase\expandafter{\romannumeral2}: Grasping Only Training.} In this stage, the parameters of the pre-trained PushNet $\phi_p$ are fixed,  and we mainly train the GraspNet $\phi_g$ with a relatively scattered scene to simulate the scenarios where objects have already been well singulated. Inspired by \cite{zeng2018learning}, we conduct object-agnostic grasping tasks and use the reward function as follows:

\begin{equation}
    R_g=\begin{cases}1.5, &\text{if grasping successfully} \\0, &\text{otherwise}
    \end{cases}
\end{equation}

Since the previous stage has created enough space for object-agnostic grasping, training a grasp-only policy maximizes the grasping success thanks to the stage \uppercase\expandafter{\romannumeral1}.

\textbf{Stage \uppercase\expandafter{\romannumeral3}: Coordination.} In this stage, we combine the stage \uppercase\expandafter{\romannumeral1} and \uppercase\expandafter{\romannumeral2} as the SaG policy $\pi$ for pushing and grasping collaboration. The pushing action is executed iteratively until the graph density value $d(G)$ reaches zero or grasp trials reaches the maximum pushing number, while the grasping action dominates when the objects are well singulated.

Algorithm 1 summarizes the details of the SaG policy learning in Supplementary Section A.1, and the training and implementation details can be found at Supplementary Section A.2.

\subsection{Mask Generation}
Through SAG interactions, we self-generate object annotations to be used to improve the segmentation model. 
Prior work \cite{eitel2019self} has explored the similar idea, but it cannot filter out multi-object moving cases as it often generates inaccurate training labels that negatively affects the transfer learning. 

We propose a learning-based binary classifier to identify single object moving cases using optical flow and apply a motion cue post-processing method on them. The classifier takes optical flow and task relevant features as input and outputs the single object moving probability. The task relevant features consist of graph density $d(G)$, average and variance of pairwise center location distance $a_d$ and $a_{var}$, and target border occupancy ratio $r_b$ as defined in \cite{yang2020deep}. Since the training data for the flow classifier is collected from simulation only and the real robot setting has a domain gap from simulation, we consider such task relevant features to help the classification. We obtain object's ground truth location directly from the simulation (V-REP~\cite{rohmer2013v}) and by comparing the object locations change. We set the probability of single object movement to be $1$ as the ground truth label when only one object was moving and $0$ otherwise.

We compute the optical flow using the FlowNet2 \cite{ilg2017flownet} with images $I_t$ and $I_{t+1}$ before and after executing the motion primitive $a_t$, respectively, and feed the optical flow together with task relevant features to the classifier. Inspired by~\cite{eitel2019self}, we use normalized graph cut on each optical flow to obtain a set of segments in binary mask format $L_t = \{ l_t^{1},..., l_t^{N}\}$ for frame $I_t$, and we select segment $l_t^{n} \in L_t$ that satisfies related constraints (e.g., location, size). We add the RGB image $I_t$ and its corresponding binary mask $l_t^{n}$ as a ground truth label into the training dataset $D = \{(I_0,l_0^{n_0}),...,(I_t,l_t^{n_t})\}$ for transfer learning.

\subsection{Mask R-CNN Transfer Learning}
We use the Mask R-CNN \cite{he2017mask} model pre-trained on COCO instance segmentation dataset with the ResNet-50-FPN backbone implemented by Detectron2 \cite{wu2019detectron2}. It is a common practice to use a baseline model pre-trained on a well-annotated standard image dataset, for instance ImageNet \cite{russakovsky2015imagenet}, where the backbone serves as a universal feature extractor in the network. Moreover, the Feature Pyramid Network (FPN) \cite{he2017mask} type backbone extracts image features from different scales, which provides better anchors prediction in various levels. 
We fine-tune the segmentation model with the self-generated dataset $D$. 
The details of segmentation results can be found at Section \ref{experiments}.

%%%%%%%%%%%%%%%%%%%%%%%%%%%% supplementary 2 %%%%%%%%%%%%%%%%%%%%%%%%%

\section{Experiments} \label{experiments}

In this section, we conduct multiple experiments to evaluate the proposed \textbf{SaG} approach. The goals of the experiments are $1)$ to compare our \textbf{SaG} with several baselines in both singulation and segmentation performances and $2)$ to show whether the fine-tuned \textbf{SaG} segmentation model is effective and further applicable to in other downstream robot manipulation tasks (e.g., grasping).

\subsection{Datasets and Evaluation Metrics}
\noindent\textbf{Singulation} We evaluate our singulation performance in simulation with 6 basic shape toy blocks. We conduct 200 test trials with various object arrangements for the singulation task and record $d(G)$ values. A trial is considered to be successful when the pairwise distances between all objects are above a threshold $p$ and $d(G)$ reaches zero within $8$ pushes.

\noindent\textbf{Segmentation} We collect 404 and 200 testing images for toy blocks and YCB \cite{ycb} objects in simulation. For default test setting, we randomly drop 10 toy blocks and 8 YCB objects in the workspace. Additionally, for cluttered test setting, we increase the number of objects to 18 toy blocks and 15 YCB objects where all objects are located in a dense pile. For real robot default testing, we manually labeled 100 images with 6 to 8 objects in the workspace. In addition, 50 images are labeled with up to 16 objects in a clutter for the cluttered test cases.

We evaluate the instance segmentation performance with the standard MS COCO evaluation metric,  average precision (AP). We also use another evaluation metric as defined in \cite{Dave_2019_ICCV} to compare current state-of-the-art non-interactive segmentation method, where scores for segmentation instances are not provided and cannot be evaluated with COCO AP. To compute the overlap precision, recall, and F-measure (P/R/F), the Hungarian matching method is used for the predicted and ground truth masks. Given the matching, the P/R/F are computed by $P = \frac{\sum_i \left|a_i \cap g(a_i) \right|}{\sum_i \left|a_i\right|}$, $R = \frac{\sum_i \left|a_i \cap g(a_i) \right|}{\sum_j \left|g_j\right|}$, $F = \frac{2PR}{P+R}$, where $a_i$ denotes the set of pixels belonging to predicted object $i$, $g(a_i)$ is the ground truth matched to each predicted region, and $g_j$ denotes ground truth pixels of object $j$.

\subsection{Singulation Performance}
We utilize the simulation environment in V-REP \cite{rohmer2013v} running a UR5 arm with an RG2 gripper.
Five baselines are compared with our approach: $1)$ \textbf{SaG-{maskall}}, the baseline without the most cluttered object mask projection map $m_t$ and filters the Q maps by the binary mask projection map $h_t$, $2)$ \textbf{SaG-{maskone}}, the baseline without the $m_t$ input and filters the Q maps with $m_t$, $3)$ \textbf{VPG}, target agnostic pushing and grasping to clean the cluttered objects \cite{zeng2018learning}. $4)$ \textbf{SaG no \boldmath$m_t$}, our proposed method without $m_t$, and $5)$ \textbf{SaG no \boldmath$h_t$}, our proposed method without $h_t$.

\begin{figure}[t]
    \centering
    \begin{minipage}{.47\textwidth}
    \centering
    \includegraphics[width=1\linewidth]{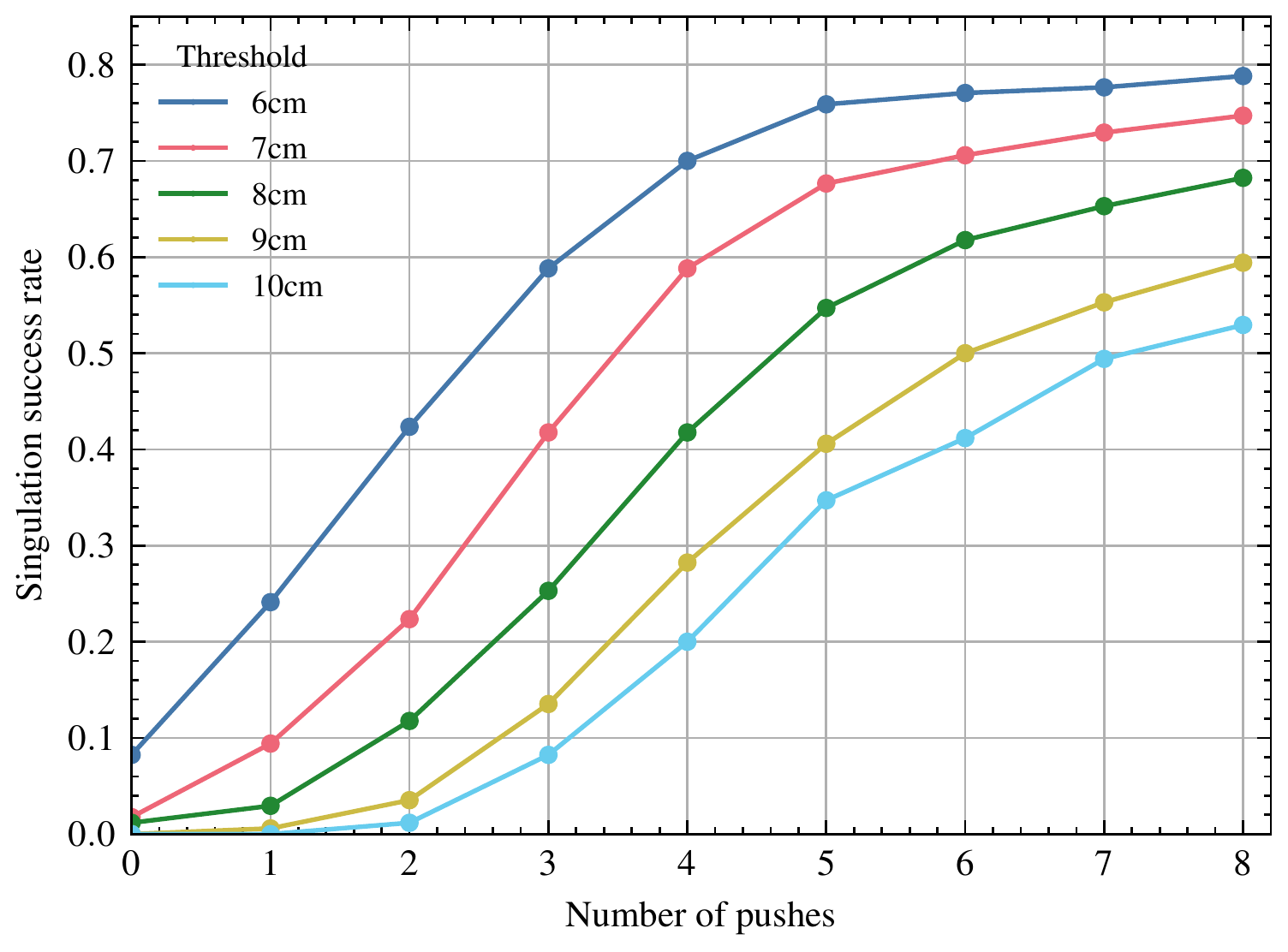}
    \caption{Singulation policy performance with different distance thresholds.}
    \label{diff_thresh}
    \end{minipage}%
    \quad
    \begin{minipage}{0.47\textwidth}
    \centering
    \includegraphics[width=1\linewidth]{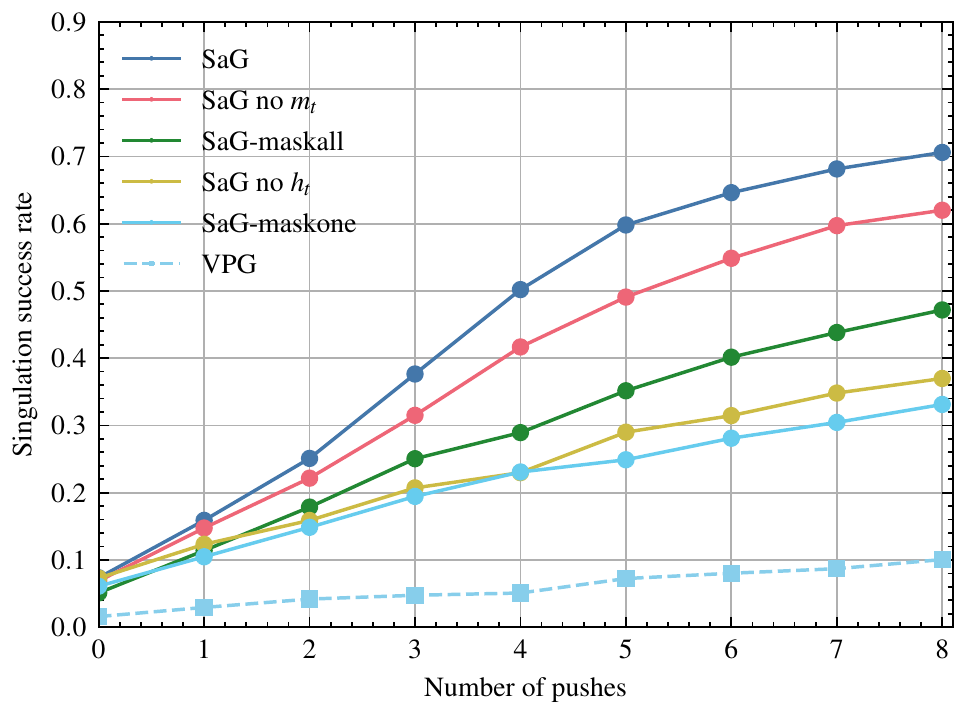}
    \caption{Singulation success rate for different baselines.}
    \label{bs}
    \end{minipage}
\end{figure}

Fig. \ref{diff_thresh} shows the average singulation success rate with different pairwise distance thresholds $p$ from $6cm$ to $10cm$ versus the number of pushes. The pushing motions achieve about 80$\%$ singulation success rate with the small threshold of $6cm$ and over 50$\%$ with the large threshold of $10cm$ after eight pushes. 

To show that \textbf{SaG} is effective, we further conduct the five push-only baseline comparisons mentioned above. We reuse the test cases when evaluating \textbf{SaG} singulation success for each method. The average singulation success rate for each baseline combines performance measurement with thresholds from $6cm$ to $10cm$. Fig. \ref{bs} demonstrates that the \textbf{VPG} push-only method barely has the object singulation effect. On the other hand, our proposed approach improves the performance by a large margin about $60\%$ after eight pushes. Although the singulation task is target-agnostic, the results of \textbf{SaG no \boldmath$m_t$} and \textbf{SaG no \boldmath$h_t$} show that providing the network with global and local clutter information helps improve the overall performance. \textbf{SaG-maskall} may push objects that have already been well singulated, resulting in an ineffective pushing policy. While \textbf{SaG-maskone} always pushes the most cluttered mask, it lacks the global object arrangement information, resulting in unsatisfactory performance as well.

\begin{table}[ht]
\centering
\caption{Segmentation results in simulation with toy blocks. Ablation study on different architectures of SaG pipeline.}
    \begin{tabular}{|c|c|ccc|ccc|}
       	\hline
	  	\multirow{2}{*}{Method} &\multirow{2}{*}{$OPF$} & \multicolumn{3}{c|}{Default} & \multicolumn{3}{c|}{Cluttered} \\
        & & {$AP_{50}$} & {$AP_{75}$} & {$AP_{50:95}$} & {$AP_{50}$} & {$AP_{75}$} & {$AP_{50:95}$} \\ 
      	\hline
      	SaG (Ours)&\checkmark&\textbf{98.7} &\textbf{96.6}&\textbf{87.8}&\textbf{88.6}&\textbf{81.0}&\textbf{73.0}\\
      	SaG (Ours)& &96.8&94.7&81.5&87.8&80.6&72.1\\
  		SaG no $m_t$&\checkmark&98.3&95.9&81.6&86.1&78.2&69.1\\
  		SaG no $h_t$&\checkmark&93.2&90.5&79.0&82.6&73.1&65.4\\
  		SaG grasp&\checkmark&97.1&94.5&83.4&86.8&79.7&71.7\\
  		SaG push&\checkmark&96.7&94.1&80.1&85.7&77.0&69.7\\
  		SaG push& &94.2&92.0&77.7&85.5&76.9&68.2\\
  		SaG-maskone&\checkmark&98.0&95.7&86.0&87.3&77.9&70.5\\
  		SaG-maskall&\checkmark&97.9&95.7&86.0&85.6&77.1&69.5\\
  		VPG\cite{zeng2018learning}&\checkmark & 89.2&86.3&69.4&81.4&72.1&65.0\\
  		SelfDeepMask \cite{eitel2019self}&\checkmark&77.4&66.4&53.7&52.0&32.0&29.9\\
  		SelfDeepMask \cite{eitel2019self}& &74.6&62.1&50.4&43.9&26.9&24.6\\
  		DeepMask \cite{pinheiro2015learning}& \checkmark&71.9&48.0&41.1&50.4&31.4&29.3\\
  		SBI \cite{pathak2018learning}& &72.1&54.2&45.6&52.8&29.1&27.3\\
   		\hline
    \end{tabular}
    \label{tab:ablation}
\end{table}

\begin{figure}[t]
\centering
\includegraphics[width=1\textwidth]{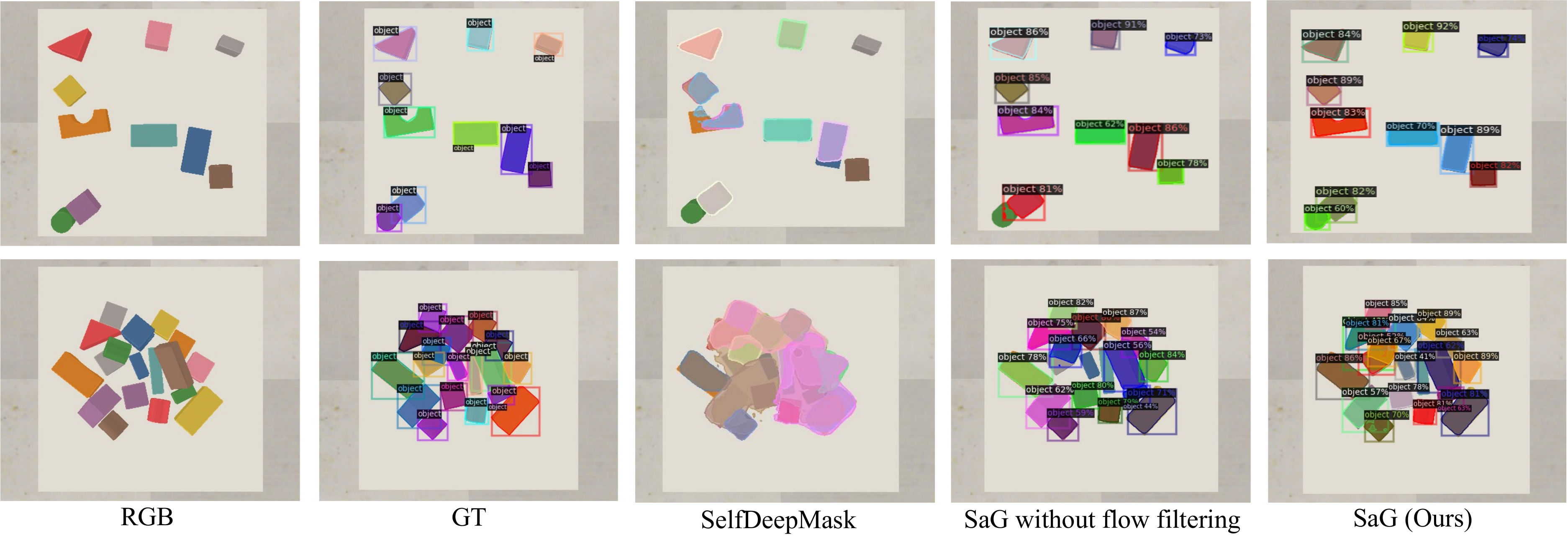}
\caption{Basic toy blocks segmentation qualitative results. The top row is related to a default test case and the bottom row is the highly cluttered test case.}
\label{sim_exp_visualization} 
\end{figure}

\subsection{Interactive Segmentation} 

We evaluate the instance segmentation performance with transfer learning in simulation. The Detectron2 COCO instance segmentation model with the ResNet-50-FPN backbone~\cite{wu2019detectron2} is used in our experiments. We fine-tune the model for 200, 250, and 150 iterations with 2000 toy blocks interactions, 2000 YCB objects interactions, and 1000 real robot novel objects interactions. Our models are trained with the initial learning rate of 0.0005 with SGD for optimization. Weight decay and momentum are set as 0.0001 and 0.9. All our models are trained on a single NVIDIA RTX 2080 Ti.

Note that the push proposal network in \textbf{SelfDeepMask} \cite{eitel2019self} and the complete data collection pipeline of \textbf{Seg-by-Interaction (SBI)} \cite{pathak2018learning} are not released. We instead prepared the training data with our SaG policy and fine-tuned their corresponding segmentation models. While the baselines comparison in such a way can be slightly unfair since we could not use their data collection methods, we followed their hyperparameter setting and the loss function selections. For \textbf{DeepMask} \cite{pinheiro2015learning} method, we fine-tune the pre-trained ResNet-50 $DeepMask$ model for 10 epochs.

\noindent\textbf{Toy Blocks Segmentation} The quantitative results are in Table \ref{tab:ablation}. We use the standard COCO instance segmentation average precision (AP) for segmentation evaluation. In Table \ref{tab:ablation}, $OPF$ denotes the use of optical flow filtering classifier. 
Our proposed approach combining pushing and grasping interactions provides the optimal performance of $87.8\%$ in default setting and $73.0\%$ in highly cluttered setting, both in $AP_{50:95}$. The \textbf{SaG push} method has relatively low performance since the singulation policy often moves multiple objects simultaneously, creating noisy labels. Our approach outperforms the compared baselines~\cite{zeng2018learning, eitel2019self, pinheiro2015learning, pathak2018learning} by large margins. Fig. \ref{sim_exp_visualization} shows the visualization results.

\begin{table}[ht]
\centering
\caption{Segmentation results in simulation with YCB objects \cite{ycb}.}
    \begin{tabular}{|c|c|ccc|ccc|}
       	\hline
	  	\multirow{2}{*}{Method} &\multirow{2}{*}{$OPF$} & \multicolumn{3}{c|}{Default} & \multicolumn{3}{c|}{Cluttered} \\
        & & {$AP_{50}$} & {$AP_{75}$} & {$AP_{50:95}$} & {$AP_{50}$} & {$AP_{75}$} & {$AP_{50:95}$} \\ 
      	\hline
      	SaG (Ours)&\checkmark&\textbf{92.9} &\textbf{82.8}&\textbf{73.9}&\textbf{84.7}&\textbf{66.8}&\textbf{61.7}\\
      	SaG (Ours)& &91.7&82.7&73.1&81.5&65.1&58.7\\
  		SelfDeepMask \cite{eitel2019self}&\checkmark&72.0&37.8&38.7&52.6&22.8&25.6\\
  		SelfDeepMask \cite{eitel2019self}& &70.4&39.1&38.6&51.6&23.3&25.6\\
  		DeepMask \cite{pinheiro2015learning}& \checkmark&69.8&28.2&33.6&51.3&18.9&23.6\\
  		SBI \cite{pathak2018learning}& &68.3&26.5&32.2&47.7&14.9&20.8\\
   		\hline
    \end{tabular}
    \label{tab:ycb}
\end{table}

\begin{figure}[ht]
\centering
\includegraphics[width=1\textwidth]{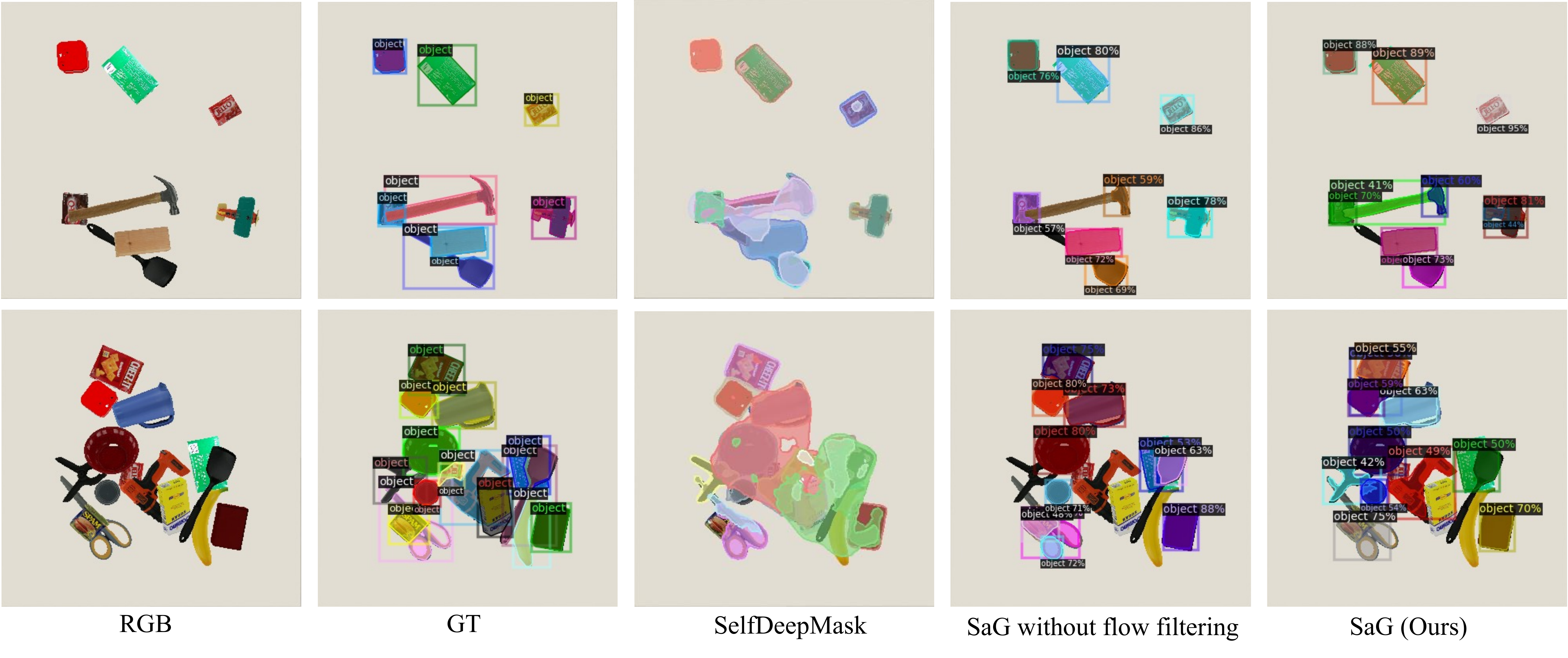}
\caption{YCB objects segmentation qualitative results. The top row shows default test setting and bottom row is the cluttered test case.}
\label{ycb_sim}
\end{figure}

\noindent\textbf{YCB Object Segmentation} We also compare baselines with more challenging objects in heavy clutter. The quantitative results are in Table \ref{tab:ycb}. Our method achieves $73.9\%$ and $61.7\%$ $AP_{50:90}$ in default and cluttered test sets, respectively, and outperforms other baselines by large margins. Segmentation visualization can be found in Fig. \ref{ycb_sim}.

\subsection{SaG Downstream Robot Task Application}
We evaluate the robotic top-down grasping task as one of the downstream tasks in simulation. The grasping performance is defined as the grasping success rate (\%) over the last 1000 attempts. 
We compare the grasping success rate with VPG \cite{zeng2018learning} (no segmentation input at all), UOIS-VPG, and SaG-VPG, where UOIS-VPG uses UOIS \cite{9382336} as the segmentation model and SaG-VPG uses the fine-tuned Mask-RCNN to provide object binary masks information as an additional input. All three methods execute the grasping actions only and were trained from scratch for 1000 epochs with 6 toy blocks. The experiment results in Table \ref{ablation2} show that the SaG based segmentation model improves the grasping performance more effectively.

\begin{table}[ht]
\centering
    \caption{Top-down grasping success rate in simulation}
    \begin{tabular}{|c|c|c|c|}
       	\hline
	  	 Settings & SaG-VPG  & UOIS-VPG & VPG \\
      	\hline
      	6 toy blocks&\textbf{85.5}&78.9&70.7\\
      	10 toy blocks&\textbf{78.4}&73.8&61.3\\
   		\hline
    \end{tabular}
    \label{ablation2}
\end{table}

\begin{table}[ht]
\centering
\caption{Segmentation results on real robot.}
    \begin{tabular}{|c|c|ccc|ccc|}
       	\hline
	  	\multirow{2}{*}{Method} &\multirow{2}{*}{$OPF$} & \multicolumn{3}{c|}{Default} & \multicolumn{3}{c|}{Cluttered} \\
        & & {$AP_{50}$} & {$AP_{75}$} & {$AP_{50:95}$} & {$AP_{50}$} & {$AP_{75}$} & {$AP_{50:95}$} \\ 
      	\hline
      	SaG (Ours)&\checkmark&\textbf{94.6}&\textbf{87.0}&\textbf{69.3}&84.3&\textbf{78.4}&\textbf{63.2}\\
      	SaG (Ours)& &88.1&80.0&61.1&\textbf{85.3}&78.3&62.5\\
  		SelfDeepMask \cite{eitel2019self}&\checkmark&79.9&57.7&51.3&70.7&51.9&43.8\\
  		SelfDeepMask \cite{eitel2019self}& &74.1&53.3&47.1&69.3&50.6&43.1\\
  		DeepMask \cite{pinheiro2015learning}& \checkmark&72.8&39.5&38.8&66.5&50.0&39.6\\
   		\hline
    \end{tabular}
    \label{tab:real_table}
\end{table}

\begin{table}[h]
\centering
\caption{SOTA comparison with the non-interactive approach on highly cluttered unseen object segmentation.}
\begin{tabular}{|c|ccc|ccc|}
        \hline
        \multirow{2}{*}{Method} & \multicolumn{3}{c|}{Overlap} & \multicolumn{3}{c|}{Boundary} \\
        & \textcolor{orange}{P} & \textcolor{cyan}{R} & \textcolor{purple}{F} & \textcolor{orange}{P} & \textcolor{cyan}{R} & \textcolor{purple}{F} \\ 
        \hline
        SaG (Ours) & \textbf{91.4} & \textbf{89.5} & \textbf{90.4} & \textbf{79.3} & \textbf{81.2} & \textbf{80.1} \\
        UOIS \cite{9382336} & 70.8 & 76.7 & 73.6 & 38.5 & 73.6 & 50.3\\
        \hline
    \end{tabular}
    \label{tab:uois}
\end{table}

\subsection{Real Robot Experiments}

\begin{figure}[t]
\centering
\includegraphics[width=0.53\textwidth]{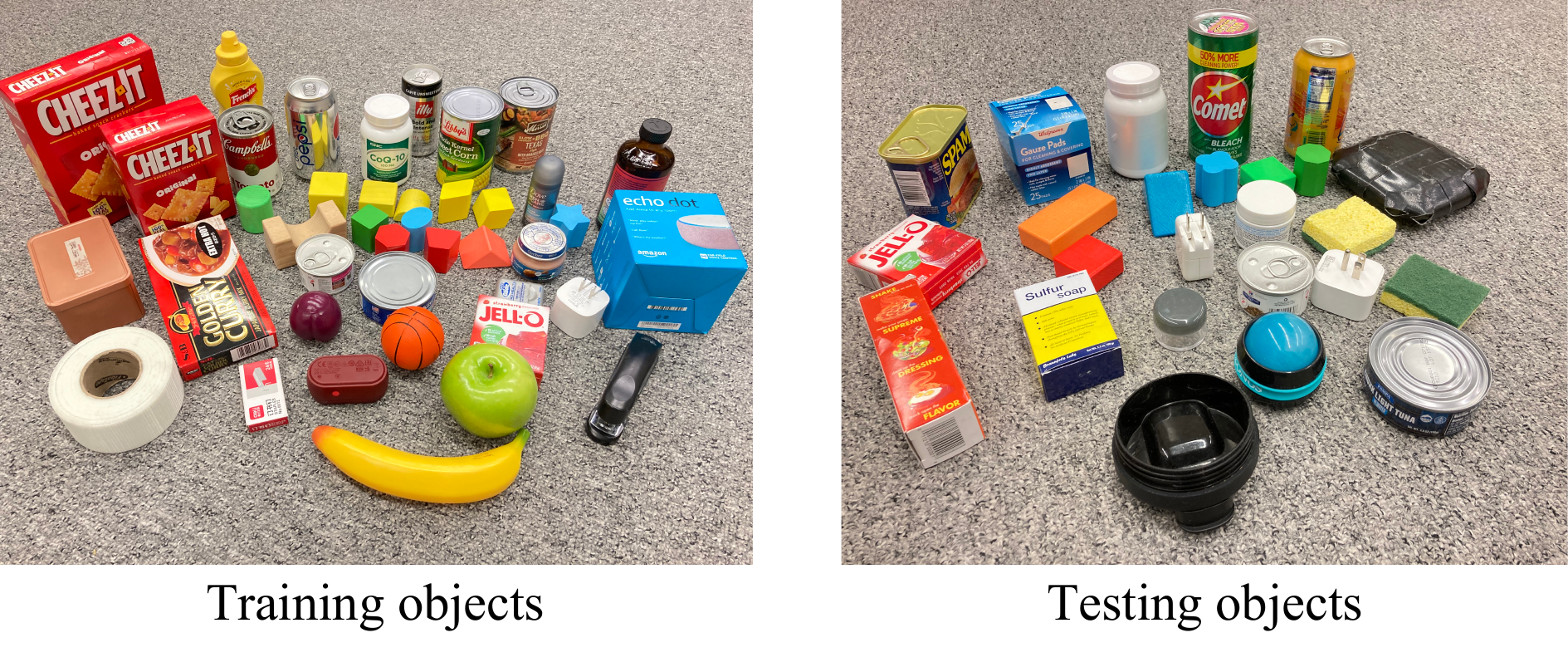}
\caption{Training and default testing object sets in real-robot experiments. The testing objects are never seen during training process.}
\label{train_test}
\end{figure}

\begin{figure}[h]
\centering
\includegraphics[width=0.96\textwidth]{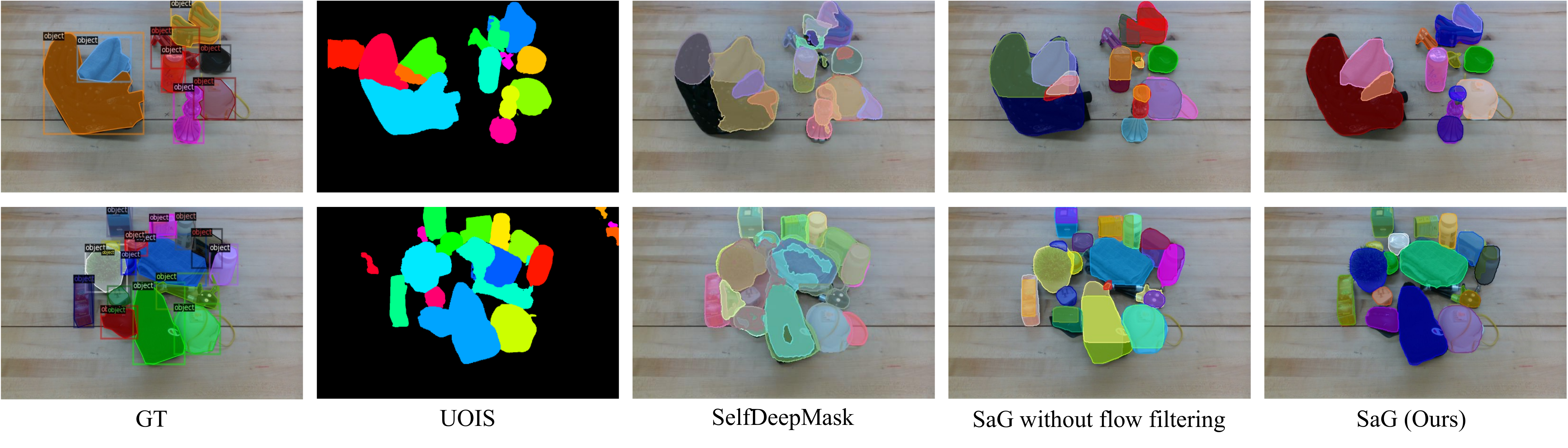}
\caption{Highly cluttered test set visualization in real robot experiments.}
\label{uois_exp}
\end{figure}

\begin{figure}[ht]
\centering
\includegraphics[width=0.96\textwidth]{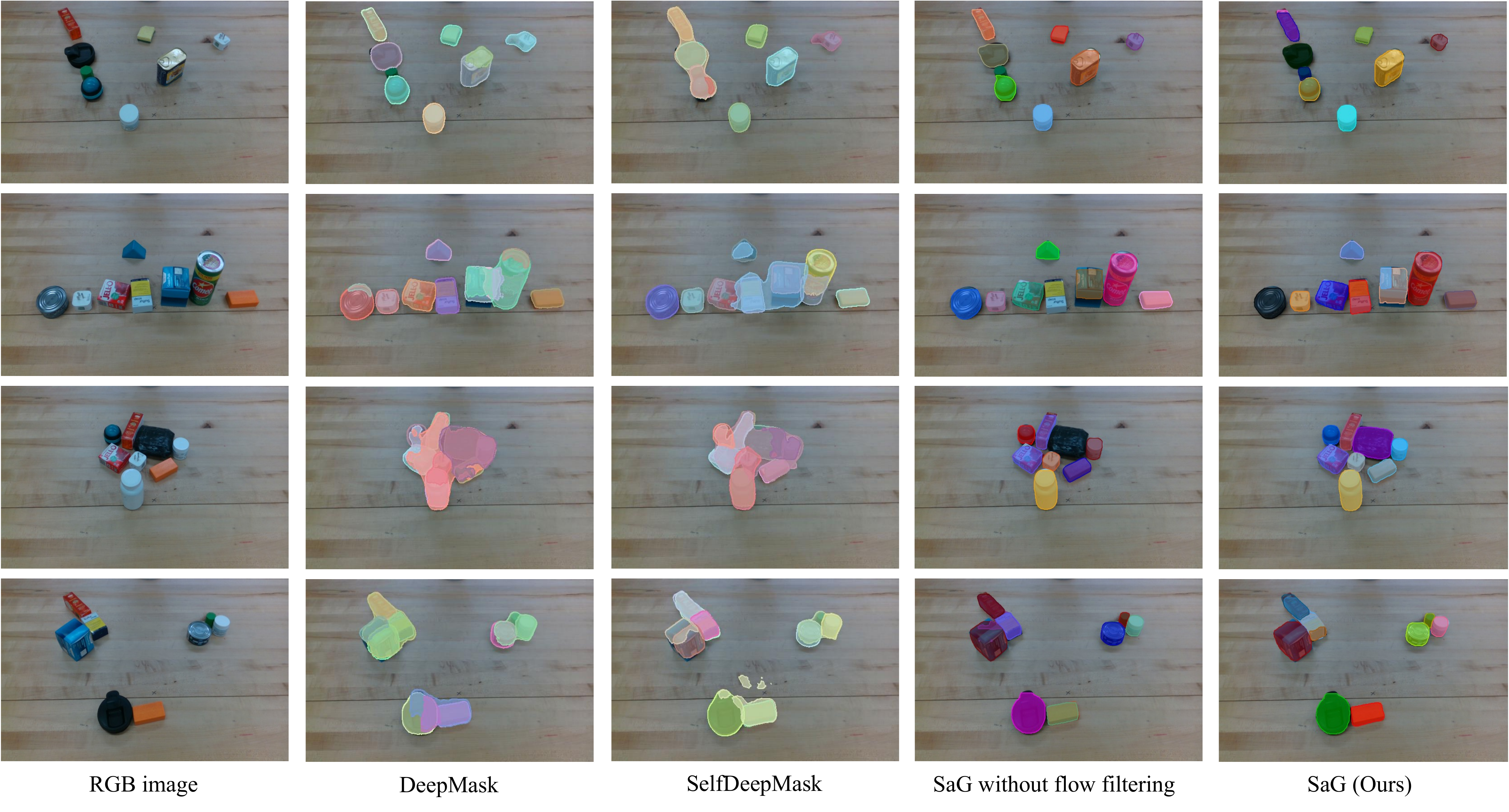}
\caption{Visualizations of test cases (sparsely distributed, a cluttered scene, and piles of objects).}
\label{real_exp_visualization}
\end{figure}

\begin{figure}[ht]
\centering
\includegraphics[width=0.95\textwidth]{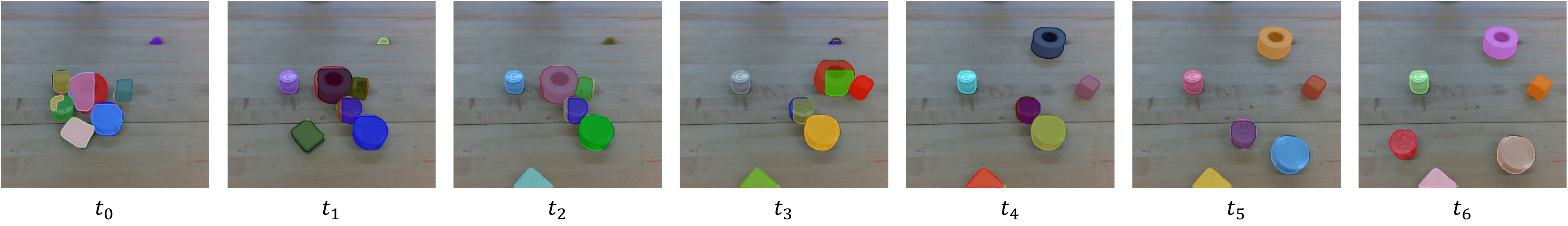}
\caption{Instance segmentation visualization with increasing number of pushes. Our singulation policy helps improve the segmentation results by breaking the clutter.}
\label{push_seg}
\end{figure}

We collect the training data via our SaG pipeline with a Franka Emika Panda robot. The \textbf{SaG} policy is only trained in simulation and not fine-tuned in real-robot setting. There are 41 different objects in the training set and 25 novel objects in the default testing set as shown in Fig. \ref{train_test}.

The quantitative segmentation results with different AP thresholds are in Table \ref{tab:real_table}. Our approach outperforms \textbf{SelfDeepMask} and \textbf{DeepMask} by $14.7$ and $21.8$ average precision points ($AP_{50}$) on default test cases, respectively. We also compare our method with pre-trained UOIS \cite{9382336} non-interactive approach in Table \ref{tab:uois}.

The qualitative visualizations are in Fig. \ref{uois_exp} and Fig. \ref{real_exp_visualization}. The results show that our segmentation model can be generalized to novel objects. Even in a cluttered scene, our model manages to segment individual objects. Fig. \ref{push_seg} shows the interactive segmentation in a push experiment, where the singulation motion separating objects further contributes to the better segmentation performance.

\section{Conclusions}

We presented an interactive object segmentation method through the SaG policy learning in the end-to-end deep Q-learning. The robot interacted with unseen objects using pushing and grasping actions and automatically generated pseudo ground truth annotations for further transfer learning. We showed our approach outperforms all the compared baselines by large margins in both simulation and real-robot experiments. Additionally, the proposed approach was applied to a downstream robot manipulation task, object grasping.

% The limitation of our work suggests two possible directions for future work. 
Although effective, the current optical-flow based classifier lowers the data collection efficiency. A future direction would learn a motion grouping method that directly provides multiple ground truth masks from optical flow. 
% The other direction would provide a singulation policy that moves one object at a time while improve the grasping success rate.

\noindent\textbf{Acknowledgements}: This work was supported in part by the Sony Research Award Program and NSF Award 2143730.

\clearpage
\bibliographystyle{splncs04}
\bibliography{egbib}

\begin{thebibliography}{10}
\providecommand{\url}[1]{\texttt{#1}}
\providecommand{\urlprefix}{URL }
\providecommand{\doi}[1]{https://doi.org/#1}

\bibitem{badrinarayanan2017segnet}
Badrinarayanan, V., Kendall, A., Cipolla, R.: Segnet: A deep convolutional
  encoder-decoder architecture for image segmentation. IEEE transactions on
  pattern analysis and machine intelligence  \textbf{39}(12),  2481--2495
  (2017)

\bibitem{boerdijk2020self}
Boerdijk, W., Sundermeyer, M., Durner, M., Triebel, R.: Self-supervised
  object-in-gripper segmentation from robotic motions. arXiv preprint
  arXiv:2002.04487  (2020)

\bibitem{bohg2017interactive}
Bohg, J., Hausman, K., Sankaran, B., Brock, O., Kragic, D., Schaal, S.,
  Sukhatme, G.S.: Interactive perception: Leveraging action in perception and
  perception in action. IEEE Transactions on Robotics  \textbf{33}(6),
  1273--1291 (2017)

\bibitem{byravan2017se3}
Byravan, A., Fox, D.: Se3-nets: Learning rigid body motion using deep neural
  networks. In: 2017 IEEE International Conference on Robotics and Automation
  (ICRA). pp. 173--180. IEEE (2017)

\bibitem{ycb}
Calli, B., Walsman, A., Singh, A., Srinivasa, S., Abbeel, P., Dollar, A.M.:
  Benchmarking in manipulation research: Using the yale-cmu-berkeley object and
  model set. IEEE Robotics Automation Magazine  \textbf{22}(3),  36--52 (2015).
  \doi{10.1109/MRA.2015.2448951}

\bibitem{chaudhary2016retrieving}
Chaudhary, K., Au, C.W., Chan, W.P., Nagahama, K., Yaguchi, H., Okada, K.,
  Inaba, M.: Retrieving unknown objects using robot in-the-loop based
  interactive segmentation. In: 2016 IEEE/SICE International Symposium on
  System Integration (SII). pp. 75--80. IEEE (2016)

\bibitem{chen2020combining}
Chen, Y., Ju, Z., Yang, C.: Combining reinforcement learning and rule-based
  method to manipulate objects in clutter. In: 2020 International Joint
  Conference on Neural Networks (IJCNN). pp.~1--6. IEEE (2020)

\bibitem{coleman1983estimation}
Coleman, T.F., Mor{\'e}, J.J.: Estimation of sparse {J}acobian matrices and
  graph coloring problems. SIAM journal on Numerical Analysis  \textbf{20}(1),
  187--209 (1983)

\bibitem{Dave_2019_ICCV}
Dave, A., Tokmakov, P., Ramanan, D.: Towards segmenting anything that moves.
  In: Proceedings of the IEEE/CVF International Conference on Computer Vision
  (ICCV) Workshops (Oct 2019)

\bibitem{deng2019deep}
Deng, Y., Guo, X., Wei, Y., Lu, K., Fang, B., Guo, D., Liu, H., Sun, F.: Deep
  reinforcement learning for robotic pushing and picking in cluttered
  environment. In: 2019 IEEE/RSJ International Conference on Intelligent Robots
  and Systems (IROS). pp. 619--626. IEEE (2019)

\bibitem{eitel2019self}
Eitel, A., Hauff, N., Burgard, W.: Self-supervised transfer learning for
  instance segmentation through physical interaction. In: 2019 IEEE/RSJ
  International Conference on Intelligent Robots and Systems (IROS). pp.
  4020--4026. IEEE (2019)

\bibitem{eitel2020learning}
Eitel, A., Hauff, N., Burgard, W.: Learning to singulate objects using a push
  proposal network. In: Robotics research, pp. 405--419. Springer (2020)

\bibitem{fang2018multi}
Fang, K., Bai, Y., Hinterstoisser, S., Savarese, S., Kalakrishnan, M.:
  Multi-task domain adaptation for deep learning of instance grasping from
  simulation. In: 2018 IEEE International Conference on Robotics and Automation
  (ICRA). pp. 3516--3523. IEEE (2018)

\bibitem{fitzpatrick2003first}
Fitzpatrick, P.: First contact: an active vision approach to segmentation. In:
  Proceedings 2003 IEEE/RSJ International Conference on Intelligent Robots and
  Systems (IROS 2003)(Cat. No. 03CH37453). vol.~3, pp. 2161--2166. IEEE (2003)

\bibitem{he2017mask}
He, K., Gkioxari, G., Doll{\'a}r, P., Girshick, R.: Mask r-cnn. In: Proceedings
  of the IEEE international conference on computer vision. pp. 2961--2969
  (2017)

\bibitem{he2016deep}
He, K., Zhang, X., Ren, S., Sun, J.: Deep residual learning for image
  recognition. In: Proceedings of the IEEE conference on computer vision and
  pattern recognition. pp. 770--778 (2016)

\bibitem{hermans2012guided}
Hermans, T., Rehg, J.M., Bobick, A.: Guided pushing for object singulation. In:
  2012 IEEE/RSJ International Conference on Intelligent Robots and Systems. pp.
  4783--4790. IEEE (2012)

\bibitem{huang2021dipn}
Huang, B., Han, S.D., Boularias, A., Yu, J.: Dipn: Deep interaction prediction
  network with application to clutter removal. In: 2021 IEEE International
  Conference on Robotics and Automation (ICRA). pp. 4694--4701. IEEE (2021)

\bibitem{huang2017densely}
Huang, G., Liu, Z., Van Der~Maaten, L., Weinberger, K.Q.: Densely connected
  convolutional networks. In: Proceedings of the IEEE conference on computer
  vision and pattern recognition. pp. 4700--4708 (2017)

\bibitem{ilg2017flownet}
Ilg, E., Mayer, N., Saikia, T., Keuper, M., Dosovitskiy, A., Brox, T.: Flownet
  2.0: Evolution of optical flow estimation with deep networks. In: Proceedings
  of the IEEE conference on computer vision and pattern recognition. pp.
  2462--2470 (2017)

\bibitem{ioffe2015batch}
Ioffe, S., Szegedy, C.: Batch normalization: Accelerating deep network training
  by reducing internal covariate shift. In: International conference on machine
  learning. pp. 448--456. PMLR (2015)

\bibitem{kenney2009interactive}
Kenney, J., Buckley, T., Brock, O.: Interactive segmentation for manipulation
  in unstructured environments. In: 2009 IEEE International Conference on
  Robotics and Automation. pp. 1377--1382. IEEE (2009)

\bibitem{kiatos2019robust}
Kiatos, M., Malassiotis, S.: Robust object grasping in clutter via singulation.
  In: 2019 International Conference on Robotics and Automation (ICRA). pp.
  1596--1600. IEEE (2019)

\bibitem{kurenkov2020visuomotor}
Kurenkov, A., Taglic, J., Kulkarni, R., Dominguez-Kuhne, M., Garg, A.,
  Mart{\'\i}n-Mart{\'\i}n, R., Savarese, S.: Visuomotor mechanical search:
  Learning to retrieve target objects in clutter. In: 2020 IEEE/RSJ
  International Conference on Intelligent Robots and Systems (IROS). pp.
  8408--8414. IEEE (2020)

\bibitem{kuzmivc2010object}
Kuzmi{\v{c}}, E.S., Ude, A.: Object segmentation and learning through feature
  grouping and manipulation. In: 2010 10th IEEE-RAS International Conference on
  Humanoid Robots. pp. 371--378. IEEE (2010)

\bibitem{le2017segmenting}
Le~Goff, L.K., Mukhtar, G., Le~Fur, P.H., Doncieux, S.: Segmenting objects
  through an autonomous agnostic exploration conducted by a robot. In: 2017
  First IEEE International Conference on Robotic Computing (IRC). pp. 284--291.
  IEEE (2017)

\bibitem{liang2021learning}
Liang, H., Lou, X., Yang, Y., Choi, C.: Learning visual affordances with
  target-orientated deep q-network to grasp objects by harnessing environmental
  fixtures. In: 2021 IEEE International Conference on Robotics and Automation
  (ICRA). pp. 2562--2568. IEEE (2021)

\bibitem{lin2014microsoft}
Lin, T.Y., Maire, M., Belongie, S., Hays, J., Perona, P., Ramanan, D.,
  Doll{\'a}r, P., Zitnick, C.L.: Microsoft coco: Common objects in context. In:
  European conference on computer vision. pp. 740--755. Springer (2014)

\bibitem{Mnih15nature}
Mnih, V., Kavukcuoglu, K., Silver, D., Rusu, A.A., Veness, J., Bellemare, M.G.,
  Graves, A., Riedmiller, M., Fidjeland, A.K., Ostrovski, G.: Human-level
  control through deep reinforcement learning. Nature  \textbf{518}(7540), ~529
  (2015)

\bibitem{nair2010rectified}
Nair, V., Hinton, G.E.: Rectified linear units improve restricted boltzmann
  machines. In: Icml (2010)

\bibitem{pinheiro2015learning}
O~Pinheiro, P.O., Collobert, R., Doll{\'a}r, P.: Learning to segment object
  candidates. Advances in neural information processing systems  \textbf{28}
  (2015)

\bibitem{pathak2018learning}
Pathak, D., Shentu, Y., Chen, D., Agrawal, P., Darrell, T., Levine, S., Malik,
  J.: Learning instance segmentation by interaction. In: Proceedings of the
  IEEE Conference on Computer Vision and Pattern Recognition Workshops. pp.
  2042--2045 (2018)

\bibitem{rohmer2013v}
Rohmer, E., Singh, S.P., Freese, M.: V-rep: A versatile and scalable robot
  simulation framework. In: 2013 IEEE/RSJ International Conference on
  Intelligent Robots and Systems. pp. 1321--1326. IEEE (2013)

\bibitem{russakovsky2015imagenet}
Russakovsky, O., Deng, J., Su, H., Krause, J., Satheesh, S., Ma, S., Huang, Z.,
  Karpathy, A., Khosla, A., Bernstein, M., et~al.: Imagenet large scale visual
  recognition challenge. International journal of computer vision
  \textbf{115}(3),  211--252 (2015)

\bibitem{sarantopoulos2020split}
Sarantopoulos, I., Kiatos, M., Doulgeri, Z., Malassiotis, S.: Split deep
  q-learning for robust object singulation. In: 2020 IEEE International
  Conference on Robotics and Automation (ICRA). pp. 6225--6231. IEEE (2020)

\bibitem{schiebener2014physical}
Schiebener, D., Ude, A., Asfour, T.: Physical interaction for segmentation of
  unknown textured and non-textured rigid objects. In: 2014 IEEE International
  Conference on Robotics and Automation (ICRA). pp. 4959--4966. IEEE (2014)

\bibitem{spelke1990principles}
Spelke, E.S.: Principles of object perception. Cognitive science
  \textbf{14}(1),  29--56 (1990)

\bibitem{wu2019detectron2}
Wu, Y., Kirillov, A., Massa, F., Lo, W.Y., Girshick, R.: Detectron2.
  \url{https://github.com/facebookresearch/detectron2} (2019)

\bibitem{xie2020best}
Xie, C., Xiang, Y., Mousavian, A., Fox, D.: The best of both modes: Separately
  leveraging rgb and depth for unseen object instance segmentation. In:
  Conference on robot learning. pp. 1369--1378. PMLR (2020)

\bibitem{9382336}
Xie, C., Xiang, Y., Mousavian, A., Fox, D.: Unseen object instance segmentation
  for robotic environments. IEEE Transactions on Robotics pp. 1--17 (2021)

\bibitem{xu2021efficient}
Xu, K., Yu, H., Lai, Q., Wang, Y., Xiong, R.: Efficient learning of
  goal-oriented push-grasping synergy in clutter. IEEE Robotics and Automation
  Letters  \textbf{6}(4),  6337--6344 (2021)

\bibitem{yang2020deep}
Yang, Y., Liang, H., Choi, C.: A deep learning approach to grasping the
  invisible. IEEE Robotics and Automation Letters  \textbf{5}(2),  2232--2239
  (2020)

\bibitem{zeng2020tossingbot}
Zeng, A., Song, S., Lee, J., Rodriguez, A., Funkhouser, T.: Tossingbot:
  Learning to throw arbitrary objects with residual physics. IEEE Transactions
  on Robotics  \textbf{36}(4),  1307--1319 (2020)

\bibitem{zeng2018learning}
Zeng, A., Song, S., Welker, S., Lee, J., Rodriguez, A., Funkhouser, T.:
  Learning synergies between pushing and grasping with self-supervised deep
  reinforcement learning. In: 2018 IEEE/RSJ International Conference on
  Intelligent Robots and Systems (IROS). pp. 4238--4245. IEEE (2018)

\end{thebibliography}

\end{document}